\documentclass{article}

\usepackage{amsmath,amsfonts,bm}

\def\eqref#1{equation~\ref{#1}}

\def\1{\bm{1}}

\DeclareMathAlphabet{\mathsfit}{\encodingdefault}{\sfdefault}{m}{sl}
\SetMathAlphabet{\mathsfit}{bold}{\encodingdefault}{\sfdefault}{bx}{n}

\usepackage{hyperref}
\usepackage{url}
\usepackage[utf8]{inputenc} 
\usepackage[T1]{fontenc}    
\usepackage{hyperref}       
\usepackage{url}            
\usepackage{booktabs}       
\usepackage{amsfonts}       
\usepackage{nicefrac}       
\usepackage{microtype}      
\usepackage{xcolor}         
\usepackage{makecell}
\usepackage{graphicx}
\usepackage{amsmath}
\usepackage{multirow}
\usepackage{wrapfig}
\usepackage{wrapfig}
\usepackage{float}
\usepackage{graphicx}
\usepackage{caption}
\usepackage{color}
\usepackage{colortbl}

\newcommand{\rnum}[1]{\uppercase\expandafter{\romannumeral #1\relax}}

\usepackage{microtype}
\usepackage{graphicx}
\usepackage{subfigure}
\usepackage{booktabs} 
\usepackage{paralist}

\usepackage{hyperref}

\usepackage[accepted]{icml2024}

\usepackage{amsmath}
\usepackage{amssymb}
\usepackage{mathtools}
\usepackage{amsthm}
\usepackage[inline]{enumitem}
\makeatletter
\newcommand{\inlineitem}[1][]{%
\ifnum\enit@type=\tw@
    {\descriptionlabel{#1}}
  \hspace{\labelsep}
\else
  \ifnum\enit@type=\z@
       \refstepcounter{\@listctr}\fi
    \quad\@itemlabel\hspace{\labelsep}
\fi}
\makeatother

\usepackage[capitalize,noabbrev]{cleveref}

\theoremstyle{plain}

\theoremstyle{definition}

\theoremstyle{remark}

\usepackage[textsize=tiny]{todonotes}

\expandafter\def\expandafter\normalsize\expandafter{%
    \normalsize%
    \setlength\abovedisplayskip{2pt}%
    \setlength\belowdisplayskip{4pt}%
    \setlength\abovedisplayshortskip{-4pt}%
    \setlength\belowdisplayshortskip{2pt}%
}

\icmltitlerunning{LCSM}

\begin{document}

\twocolumn[
\icmltitle{Unlocking the Secrets of Linear Complexity Sequence Model from A Unified Perspective}



\icmlsetsymbol{equal}{*}
\begin{icmlauthorlist}
\icmlauthor{Zhen Qin}{tp}
\icmlauthor{Xuyang Shen}{lab}
\icmlauthor{Dong Li}{lab}
\icmlauthor{Weigao Sun}{lab}
\icmlauthor{Stan Birchfield}{nv}
\icmlauthor{Richard Hartley}{anu}
\icmlauthor{Yiran Zhong}{lab}
\end{icmlauthorlist}

\icmlaffiliation{lab}{OpenNLPLab, Shanghai AI Lab}
\icmlaffiliation{nv}{NVIDIA}
\icmlaffiliation{anu}{The Australian National University}
\icmlaffiliation{tp}{Taptap}

\icmlcorrespondingauthor{Yiran Zhong}{zhongyiran@gmail.com}

\icmlkeywords{Linear attention, sequence parallelism, unlimited sequence length, large language model}

\vskip 0.3in
]



\printAffiliationsAndNotice{}  

\begin{abstract}
We present the Linear Complexity Sequence Model (LCSM), a comprehensive solution that unites various sequence modeling techniques with linear complexity, including linear attention, state space model, long convolution, and linear RNN, within a single framework. The goal is to enhance comprehension of these models by analyzing the impact of each component from a cohesive and streamlined viewpoint. Specifically, we segment the modeling processes of these models into three distinct stages: Expand, Oscillation, and Shrink (EOS), with each model having its own specific settings. The Expand stage involves projecting the input signal onto a high-dimensional memory state. This is followed by recursive operations performed on the memory state in the Oscillation stage. Finally, the memory state is projected back to a low-dimensional space in the Shrink stage. We perform comprehensive experiments to analyze the impact of different stage settings on language modeling and retrieval tasks. Our results show that data-driven methods are crucial for the effectiveness of the three stages in language modeling, whereas hand-crafted methods yield better performance in retrieval tasks.
\end{abstract}

\section{Introduction}

Transformers~\citep{vaswani2017attention} have revolutionized deep learning in recent years.  
Nevertheless, the escalating demand for extra-long sequence modeling and the abundance of training data have necessitated the development of sequence modeling techniques beyond the original formulation. 
In particular, recent years have witnessed the increased development of linear complexity sequence modeling techniques. Various advanced efficient approaches have emerged lately, exhibiting notable performance improvements and narrowing the performance gap with transformer-based methods~\citep{qin2023scaling,2312.00752,GLA}.


\begin{figure*}[t]
    \centering
    \includegraphics[width=1\textwidth]{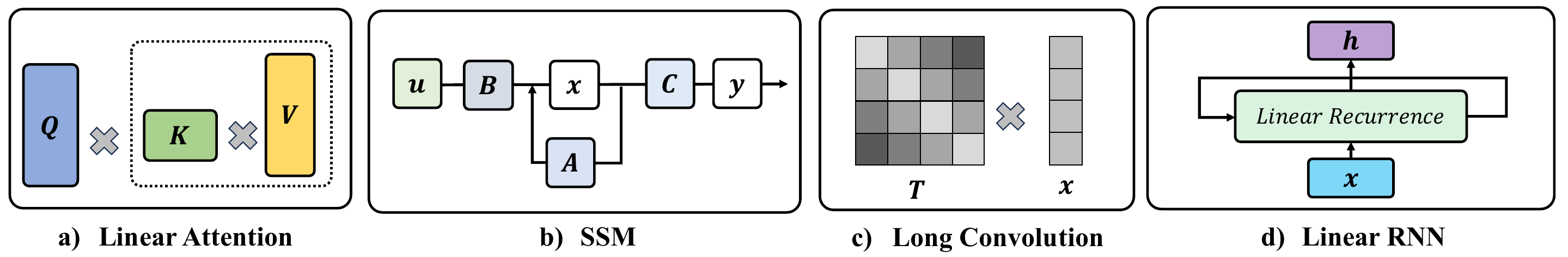}
    \vspace{-8mm}
    \caption{\textbf{Primary Groups of Linear Complexity Sequence Model.}}
    \vspace{-4mm}
    \label{fig:esm}
\end{figure*}

These methods can be categorized into four primary groups.
\emph{1) Linear attention}~\citep{xfmrsarernns,choromanski2021rethinking} involves a series of techniques aimed at computing attention matrices using the "right-product kernel trick." This approach computes the key-value production first, thereby avoiding quadratic query-key production. Recent studies~\citep{qin2023scaling,retnet} have shown that Linear attention can achieve performance levels comparable to softmax attention while significantly improving speed.
\emph{2) State Space Model}~\cite{gu2022efficiently,s5} (SSM) encompasses methods that utilize State Space Equations for sequence modeling. Through the use of special initialization, diagonalization assumptions, and mixed techniques, SSM methods can achieve performance levels comparable to softmax attention methods in language modeling~\citep{2312.00752}.
\emph{3) Long convolution}~\citep{qin2023toeplitz,hyena} models adopt the concept of CNN but utilize a kernel with a size equal to the input sequence length. Fast Fourier Transforms (FFT) are employed to reduce computational complexities to log-linear~\citep{simplelongconv,fu2023flashfftconv}. However, the language modeling performance of these methods still falls short of that of transformers.
\emph{4) Linear RNNs}~\citep{HGRN,lru} enable parallel training of the RNN, and demonstrate competitive performance against similarly scaled Transformers.

Recent research suggests that there are inherent connections between these methodologies. For instance, Long convolution models can be transformed into SSM models during inference using a closed-form solution~\citep{qin2023accelerating}. Similarly, linear attention and SSM models can conduct inference in an RNN-style fashion~\citep{xfmrsarernns}. By delving into the intrinsic connections, in this paper, we present a unified framework called Linear Complexity Sequence Model (LCSM) that effectively encompasses all established methods for linear complexity sequence modeling. This framework facilitates a deeper understanding of the operational principles underlying these methods. 

The LCSM model consists of three stages: \textbf{Expand, Oscillation, and Shrink (EOS)} as well as a Memory state. During the Expand stage, input signals are transformed into a high-dimensional memory state using a predefined projection function such as the outer product. The memory state is then combined with previous memory states through recursive operations in the Oscillation stage. Finally, the combined memory state is reduced to a low-dimensional output space in the Shrink stage. Different methodologies often exhibit variations in three key areas: 1) parameterization strategies used to calculate the Expand and Shrink parameters, 2) data dependency strategies for EOS, and 3) construction methods for the Oscillation state.

Specifically, we examine two parameterization strategies, eight EOS data dependency strategies and twelve approaches for constructing the Oscillation state in order to assess the impact of each option. Ablation studies are conducted to analyze the impact of activation functions and hyperparameters, providing complementary insights. Our main focus is on evaluating the language modeling capabilities and information retrieval abilities of these variants. 
Our empirical findings indicates that data dependency plays a crucial role in the construction of oscillation states and the parameterization of Expand and Shrink stages in language modeling tasks, while its significance is comparatively lower in retrieval tasks. Our use of carefully hand-crafted EOS parameterization has resulted in improved overall performance in retrieval tasks.

\section{Method}

This section begins with a general definition of sequence modeling and then delves into the specifics of our Linear Complexity Sequence Model (LCSM). We will also provide examples to illustrate that all current linear complexity sequence modeling methods can be considered as a special case of LCSM.

\subsection{General Definition of Sequence Modeling}
Let us consider a sequence mapping $f: \mathbb{R}^{n \times d} \to \mathbb{R}^{n \times d}$:
\begin{equation}
\small
\begin{gathered}
{\left[\begin{array}{c}
\mathbf{y}_1^{\top} \\
\vdots \\
\mathbf{y}_n^{\top}
\end{array}\right]=\mathbf{Y}=f(\mathbf{X})=f\left(\left[\begin{array}{c}
\mathbf{x}_1^{\top} \\
\vdots \\
\mathbf{x}_n^{\top}
\end{array}\right]\right)},
\end{gathered}
\end{equation}
or $ \mathbf{Y} = f(\mathbf{X})$, where $\mathbf{X},\mathbf{Y} \in \mathbb{R}^{n \times d}$, $n$ is the temporal length of the sequence, and $d$ is the feature dimensionality.
Let $\mathbf{y}_m =f\left(\mathbf{x}_1, \ldots, \mathbf{x}_n\right)_m \in \mathbb{R}^d$ be the transpose of the $m$th row of $\mathbf{Y}$.
In particular, we consider the causal mapping $f_n: \mathbb{R}^{n \times d} \to \mathbb{R}^d$:
$$
\mathbf{y}_n = f_n(\mathbf{x}_1, \ldots, \mathbf{x}_n)
\triangleq 
f(\mathbf{x}_1, \ldots, \mathbf{x}_n)_n.
$$
Language modeling is a typical example of causal mapping.
The subsequent sections of the paper will focus on causal mapping for the purpose of simplifying the explanation, but the extension to non-causal mapping is straightforward.


\subsection{Linear Complexity Sequence Modeling}

Our proposed Linear Complexity Sequence Model (LCSM) performs the mapping sequentially as follows:
\begin{align}
\mathbf{m}_t &=g_\psi(\mathbf{o}_t, \mathbf{m}_{t-1})+\mathbf{e}_t \mathbf{i}_t^{\top} \\
\mathbf{y}_t &=\mathbf{m}_t^{\top} \mathbf{s}_t,    
\end{align}
where $\mathbf e_t \in \mathbb R^k$ is the \emph{expand state}, $\mathbf o_t \in \mathbb R^{k \times p}$ is the \emph{oscillation state}, $\mathbf s_t \in \mathbb R^k$ is the \emph{shrink state}, $\mathbf i_t \in \mathbb R^d$ is the \emph{input state}, $k$ is the expand dimension, and $p$ is defined below.  

In other words, there are three stages to the information flow of the sequence modeling process.  These stages, known as Expand, Oscillation, and Shrink (EOS), are repeated for each time step $t$:
\begin{enumerate}
    \item \emph{Expand Stage:}  The \emph{input state} $\mathbf i_t$ is expanded, via the expand state $\mathbf e_t$:
\begin{equation}
    \bar{\mathbf{m}}_t = \mathbf{e}_t \mathbf{i}_t ^\top \quad \in \mathbb R^{k\times d}.
\end{equation}

    \item 
\emph{Oscillation Stage:}  The memory state $\mathbf{m}_{t}$ is updated:
$$
\mathbf{m}_{t} = g_{\psi}(\mathbf{o}_t , \mathbf{m}_{t-1}) + \bar{\mathbf{m}}_t \quad \in \mathbb R^{k\times d},
$$
where for initialization we set $ \mathbf{m}_0 = \texttt{zeros}(k,d)$, which is a $k \times d$ matrix of zeros.
    \item 
\emph{Shrink Stage:}  The \emph{output state} $ \mathbf{y}_t $ is obtained by projecting the memory state $\mathbf{m}_{t}$ back to low-dimensional space via the shrink state $\mathbf{s}_t$:
\begin{equation}    
\mathbf{y}_t = \mathbf{m}_t^{\top} \mathbf{s}_t  \quad \in \mathbb{R}^d.
\end{equation}
\end{enumerate}

\begin{figure*}[t]
    \centering
    \includegraphics[width=0.8\textwidth]{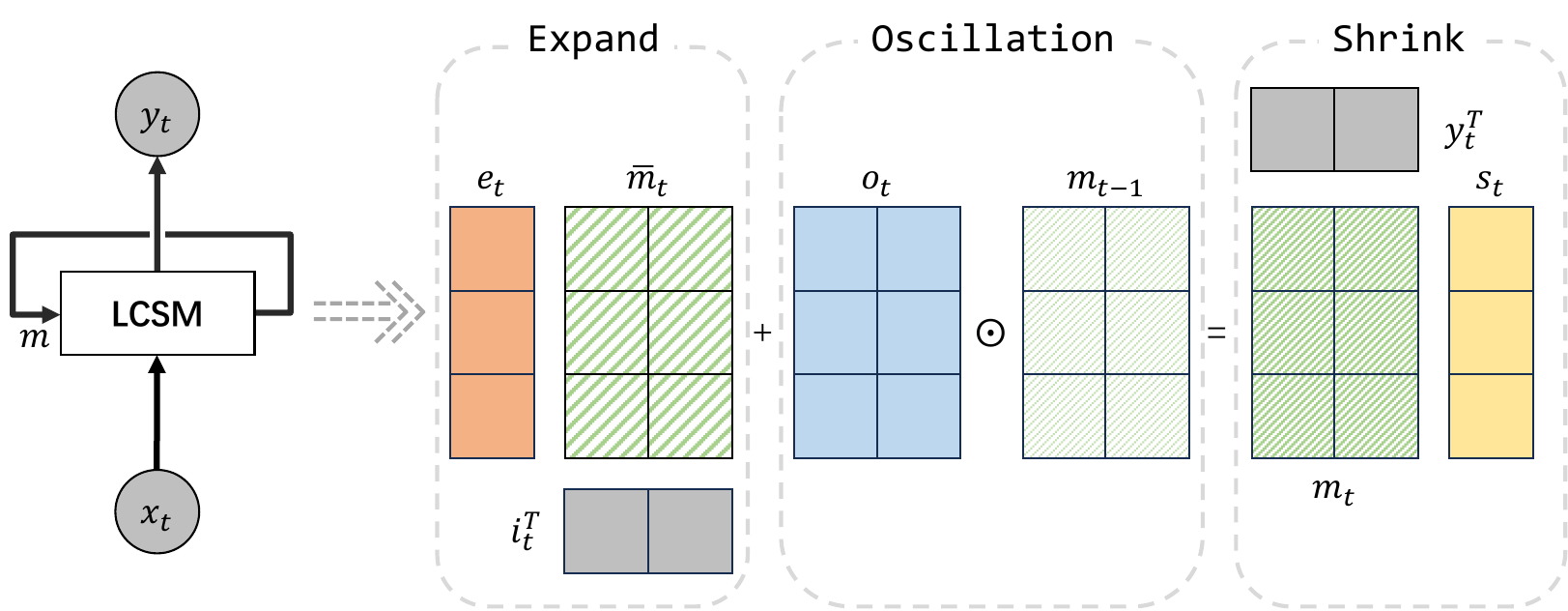}
    \caption{\textbf{Diagram of LCSM.} LCSM works in a recurrent form and consists of three states: Expand, Oscillation, and Shrink. Input signals are transformed into a high-dimensional state in the Expand stage and then recursively merged in the Oscillation stage. The combined memory state is condensed into a low-dimensional output space in the Shrink stage.}
    \label{fig:lcsm}
\end{figure*}

We consider two possible binary operators $\psi$. 
If $\psi = \odot$, the projection function is  element-wise multiplication, whereas if $\psi = \times$, it is matrix multiplication:
\begin{align}
   g_\odot(\mathbf{o}_t,\mathbf{m}_{t-1}) &= \mathbf{o}_t \odot \mathbf{m}_{t-1} \\ g_\times(\mathbf{o}_t,\mathbf{m}_{t-1}) &= \mathbf{o}_t \mathbf{m}_{t-1}.
\end{align}
From matrix dimensions, it is easy to see that
\begin{equation}
    p = \begin{cases}d & \text{if } \psi = \odot \\ k & \text{if } \psi = \times.\end{cases}
\end{equation}

The discussion is simplified by noting that, when $\mathbf o_t$ is diagonalizable, which is a common assumption in practice~\citep{gupta2022DSS}, $\psi=\times$ can be reduced to $\psi=\odot$. This is because $\mathbf o_t=\text{diag}\{{\mathbf {\bar o_ t}}\}$, where $\mathbf {\bar {o}_t}\in \mathbb R^{k}$, leads to:
\begin{equation}
\begin{aligned}
    g_\times(\mathbf o_t, \mathbf m_{t-1}) = \mathbf o_t \mathbf m_{t-1} = \text{diag}(\mathbf {\bar o}_t) \mathbf m_{t-1} \\
    = \left( \mathbf {\bar {o}}_t \mathbf 1_{\{k\}}^\top \right) \odot \mathbf m_{t-1}
    = g_\odot(\mathbf {\bar {o}}_t \mathbf 1_{\{k\}}^\top, \mathbf m_{t-1}) 
\end{aligned}  
\end{equation}
Therefore, to simplify the discussion, we only consider the case where $\psi=\odot$ in the main text, which also includes the case when $\psi=\times$ and $\mathbf o_t$ is diagonalizable.  We discuss a few examples where $\mathbf o_t$ is not diagonalizable in the Appendix~\ref{no_diag}.

\subsection{Examples}
To effectively demonstrate our unifying LCSM framework, below we describe how various well-established linear complexity sequence modeling methods can be integrated into the framework. 
The correspondence is listed in Table~\ref{table:eos}.

There are two different families of methods.
In data-dependent methods, the expand, oscillation and shrink states are either fully or partially computed based on the input $ \mathbf{x}_t $.  Notationally, this is represented using the subscript $t$, as in $\mathbf e_t$, $\mathbf o_t$, $\mathbf s_t$. Alternatively, in data-independent methods, these states are computed regardless of the input $ \mathbf{x}_t $.  Notationally, we drop the subscript in this case, leading to $\mathbf e$, $\mathbf o$, $\mathbf s$. 

 

\begin{table*}[t]
    \small
    \centering
    \caption{\textbf{Checklist for typical linear complexity sequence modeling within the defined LCSM Framework.} For each methodology, the following states and functions are outlined: Input State, Expand State, Oscillation State, Shrink State,  Memory State, and the operation. If the state is directly linked to the input sequence, the subscript $_i$ is emphasized. We use $\mathbf 1^{(k)}\in \mathbb R^k$, where $\mathbf 1^{(k)}_{j} = 1$ for $j=1,\ldots, k$, and $\mathbf J^{(kd)}=\mathbf 1^{(k)}{\mathbf 1^{(d)}}^\top \in \mathbb R^{k\times d}$.}
    \setlength{\tabcolsep}{5mm}

    \begin{tabular}{lcccccc}
    \toprule
Methods           & Input                     & Expand                          & Oscillation                                                 & Shrink                    & Memory $k\times d$  & $\psi$ \\ \midrule

S4               & $ \mathbf u_t $           & $ \mathbf B$         & $ \mathbf A$                              & $ \mathbf C$    & $k\times 1$  & $\times$  \\ 
S5               & $ \mathbf u_t$   & $ \mathbf B$         & $ \mathbf A$                              & $ \mathbf C $     & $k \times d$ & $\times$   \\ 
DSS              & $ \mathbf u_t $           & $ \mathbf B$         & $ \mathbf a \mathbf 1_k^{\top}$           & $ \mathbf C $    & $k\times d$  & $\times$   \\ 
TNN & $\mathbf x_t$ & $\mathbf{B}$ & $\mathbf A$ & $\mathbf C$ & $k\times d$ & $\times$  \\
Linear Attention & $ \mathbf v_t$ & $  \mathbf k_t$     & $ \mathbf J^{(kd)}$                       & $ \mathbf q_t$  & $k\times d$  & $\odot$   \\ 
TNL/RetNet       & $ \mathbf v_t$ & $ \mathbf k_t $      & $\lambda \mathbf J^{(k)}$                 & $ \mathbf q_t$  & $k\times d$  & $\odot$   \\ 
Mamba            & $ \mathbf u_t $   & $ \mathbf B_t$       & $ \mathbf A_t$                            & $ \mathbf C_t $   & $k\times d$  & $\odot$   \\ 
RWKV4           & $ \mathbf v_t $  & $\exp(\mathbf k_t)$ & $\exp(-w )$                              & $ \mathbf R_t $   & $1\times 1$  & $\odot$   \\ 
Cosformer        & $ \mathbf v_t $ & $ \mathbf k_t$      & $\exp(i\theta)\mathbf J^{(kd)}$           & $ \mathbf q_t$  & $k\times d$  & $\odot$   \\ 
LRPE             & $ \mathbf v_t$ & $ \mathbf k_t$      & $\exp(i \Theta) {\mathbf 1^{(d)}}^{\top}$ & $  \mathbf q_t$ & $k\times d$  & $\odot$   \\ 
GLA/GateLoop     & $ \mathbf v_t$ & $ \mathbf k_t$      & $\mathbf g_t \mathbf 1_d^{\top}$         & $  \mathbf q_t$ & $k\times d$  & $\odot$   \\ 
DUR/GFW          & $ \mathbf v_t $ & $ \mathbf k_t$      & $\mathbf g_t \bar{\mathbf  g}_t^{\top} $  & $  \mathbf q_t$ & $k\times d$  & $\odot$   \\ 
HGRN/LRN         & $\mathbf i_t$    & $1- \lambda_t$      & $ \lambda_t$                                      & $\mathbf o_t$   & $1\times 1$  & $\odot$   \\ 

        \bottomrule
    \end{tabular}
    \label{table:eos}
\end{table*}


$\bullet$ \textbf{S4}~\citep{gu2022efficiently}.  By setting $d=1$, this method 
performs channel-wise mapping.  The full mapping $f: \mathbb{R}^{n\times d} \to \mathbb{R}^{n\times d}$ requires a bank of these single-input single-output (SISO) SSMs.  In other words, the computation is repeated $d$ times, as in, $f_i$, where $i=1,\ldots,d$. 
(Note that $\mathbf u_t, \mathbf y_t \in \mathbb R^k$.  And
$\mathbf A, \mathbf B, \mathbf C $ are obtained through SSM parameterization.) 

$\bullet$ \textbf{S5}~\citep{s5}.
The recurrence equation of S5 is the same as S4, with the only difference being that the bank of SISO SSMs is replaced by a single multi-input multi-output (MIMO) SSM.  The direct definition of the mapping is $ \mathbb{R}^{n\times d} \to \mathbb{R}^{n\times d}$ and $\mathbf B, \mathbf C \in \mathbb R^{k\times d}  $.

$\bullet$ \textbf{DSS}~\citep{gupta2022DSS}.
The recurrence equation of DSS is same as S5, except that the computation is greatly simplified by diagonalizing $\mathbf A=\mathrm{Diag}\{\mathbf a\}\in \mathbf R^{k\times k}$.

$\bullet$ \textbf{TNN}~\citep{qin2023toeplitz}.
Toeplitz neural networks (TNNs) exhibit competitive performance on long-range tasks with log-linear space-time complexity.  As shown in~\citep{qin2023accelerating}, a TNN can be losslessly converted to an SSM, where $\mathbf C=\mathbf J^{(kd)}\in \mathbb R^{k\times d}, \mathbf B \in \mathbb R^{k\times d}, \mathbf A=\mathrm{Diag}\{\lambda_1, \ldots, \lambda_k\}\in \mathbb R^{k\times k}$, get $\mathbf u_t$ from $\mathbf x_t$ through linear projection, and it can be expressed as a recursive formula:
$$
\mathbf{m}_t=\mathbf{A} \mathbf{m}_{t-1}+\mathbf{B} \mathbf{u}_t^{\top},\\ \quad
\mathbf{y}_t=\mathbf{m}_t^{\top} \mathbf{C} .
$$

$\bullet$ \textbf{Linear Attention}~\citep{xfmrsarernns}.
In Linear Attention, we obtain query $ \mathbf{q}_t \in \mathbb{R}^{k} $, key $ \mathbf{k}_t \in \mathbb{R}^{k} $, value $ \mathbf{v}_t \in \mathbb{R}^{d} $ from the input $ \mathbf{x}_t \in \mathbb{R}^{d} $ through linear projection, and recursively calculate as follows:
$$
[\mathbf{kv}]_t = [\mathbf{kv}]_{t-1} + \mathbf{k}_t \mathbf{v}_t^\top, \\ \quad
\mathbf{y}_t = [\mathbf{kv}]_t^{\top} \mathbf{q}_t.
$$

$\bullet$ \textbf{TNL/RetNet}~\citep{qin2023scaling,retnet}.
TNL/RetNet is a form of Linear Attention with exponential decay, and the method for getting $\mathbf q_t, \mathbf k_t, \mathbf v_t$ is the same as those in Linear Attention, and $\lambda$ is a predefined parameter that cannot be learned. Its recursive calculation is:
$$
[\mathbf{kv}]_t =\lambda [\mathbf{kv}]_{t-1} + \mathbf{k}_t \mathbf{v}_t ^\top, \\ \quad
\mathbf{y}_t = [\mathbf{kv}]_t^{\top } \mathbf{q}_t.
$$

$\bullet$ \textbf{Mamba}~\citep{2312.00752}.
Mamba can be seen as a data-dependent S4 with elementwise multiplication. It uses the similar method to get $\mathbf u_t, \mathbf A, \mathbf B, \mathbf C$, the $\mathbf {A_t}, \mathbf B_t, \mathbf C_t$ are computed through $\mathbf x_t$ and $\mathbf A, \mathbf B, \mathbf C$. Its recurrence equation is defined as:
$$
\mathbf{m}_t=\mathbf{A}_t \odot \mathbf{m}_{t-1}+\mathbf{B}_t \mathbf{u}_t^{\top}, \\ \quad
\mathbf{y}_t=\mathbf{m}_t^{\top} \mathbf{C}_t .
$$

$\bullet$ \textbf{RWKV4}~\citep{peng2023rwkv}.
In RWKV4, we get $\mathbf r_t, \mathbf k_t, \mathbf v_t$ through linear projection from input $\mathbf x_t$ and $\mathbf w$ as a learnable weight. Ignoring the denominator of RWKV4, the recurrence equation can be simplified as:
$$
\mathbf {m}_t =\exp(-w) \mathbf {m}_{t-1} + \exp( \mathbf k_t) \mathbf  v_t^\top , \\ \quad
\mathbf y_t =  \mathbf {m}_t^{\top } \mathbf r_t.
$$
Similar to S4, RWKV4 uses channel-wise mapping $ f_i, i=1,\ldots ,d $ of $ \mathbb{R}^{n\times 1}\to \mathbb{R}^{n\times 1} $. 



$\bullet$ \textbf{Cosformer}~\citep{zhen2022cosformer}.
In Cosformer, we obtain query $ \mathbf q_t \in \mathbb{R}^{k} $, key $ \mathbf k_t \in \mathbb{R}^{k} $, value $ \mathbf v_t \in \mathbb{R}^{d} $ from the input $ \mathbf x_t \in \mathbb{R}^{d} $ and a prefined $\theta$ (not learnable). Then recursively calculate as follows:
$$
[\mathbf {kv}]_t =\exp(i\theta)[\mathbf {kv}]_{t-1} + \mathbf k_t \mathbf  v_t^\top, \\ \quad
\mathbf y_t =  \mathrm{Rel}\{[\mathbf {kv}]_t \}^{\top } \mathbf q_t.
$$
We provide the proof in the Appendix~\ref{proof}.

$\bullet$ \textbf{LRPE}~\citep{qin2023linearized}.
In LRPE, we obtain query $ \mathbf q_t \in \mathbb{R}^{k} $, key $ \mathbf k_t \in \mathbb{R}^{k} $, value $ \mathbf v_t \in \mathbb{R}^{d} $ from the input $ \mathbf x_t \in \mathbb{R}^{d} $, $\theta$ as a learnable weight and recursively calculate as follows:
\begin{equation}
    \begin{aligned}
        [\mathbf {kv}]_t &=\Lambda [\mathbf {kv}]_{t-1} + \mathbf k_t \mathbf  v_t^\top , \\
\Lambda &=\mathrm{diag}\{\exp(i\theta_1),\ldots, \exp(i\theta_k) \},   \\
\mathbf y_t &=  \mathrm{Rel}\{[\mathbf {kv}]_t \}^{\top } \mathbf q_t.
    \end{aligned}
\end{equation}
We provide the proof in the Appendix~\ref{proof}.

$\bullet$ \textbf{GLA/GateLoop}~\citep{GLA,katsch2024gateloop}.
In GLA/GateLoop, we obtain query $ \mathbf q_t \in \mathbb{R}^{k} $, key $ \mathbf k_t\in \mathbb{R}^{k} $, value $ \mathbf v_t \in \mathbb{R}^{d} $, decay $ \mathbf g_t\in \mathbf R^k $ from the input $ \mathbf x_t \in \mathbb{R}^{d} $ and recursively calculate as follows:
$$
[\mathbf{kv}]_t =\mathrm{Diag}\{\mathbf g_t \} [\mathbf{kv}]_{t-1} + \mathbf{k}_t \mathbf{v}_t ^\top, \quad
\mathbf{y}_t = [\mathbf{kv}]_t^{\top } \mathbf{q}_t.
$$

$\bullet$ \textbf{DUR/GFW}~\citep{2210.04243,schlag2018gated}
In DUR/GFW, we obtain query $ \mathbf q_t \in \mathbb{R}^{k} $, key $ \mathbf k_t\in \mathbb{R}^{k} $, value $ \mathbf v_t \in \mathbb{R}^{d} $, decay $ \mathbf g_t\in \mathbf R^k, \bar{\mathbf g}_t \in \mathbf R^d $ from the input $ \mathbf x_t \in \mathbb{R}^{d} $, and recursively calculate as follows:
$$
[\mathbf{kv}]_t =(\mathbf g_t \bar{\mathbf g}_t^\top) \odot [\mathbf{kv}]_{t-1} + \mathbf{k}_t \mathbf{v}_t ^\top, \quad
\mathbf{y}_t = [\mathbf{kv}]_t^{\top } \mathbf{q}_t.
$$

$\bullet$ \textbf{HGRN/LRN}~\citep{HGRN,martin2018parallelizing}.
In HGRN/LRN, we obtain output gate $ \mathbf o_t \in \mathbb{R}^{1} $, forget gate $ \mathbf f_t\in \mathbb{R}^{1} $, input state $ \mathbf i_t \in \mathbb{R}^{1} $ from the input $ \mathbf x_t \in \mathbb{R}^{1} $, and recursively calculate as follows:
$$
\mathbf{h}_t =\mathbf f_t \odot \mathbf{h}_{t-1} + (1-\mathbf{f}_t) \mathbf{i}_t ^\top, \quad
\mathbf{y}_t = \mathbf{h}_t^\top  \mathbf{o}_t.
$$
Similar to S4,  HGRN/LRN uses channel-wise mapping $ f_i,i=1,\ldots ,d $ of $ \mathbb{R}^{n\times 1}\to \mathbb{R}^{n\times 1} $. 


\subsection{State Calculation}
In the preceding section, it was demonstrated that various sequence modeling methods employ different approaches to calculate the EOS states. Here, we summarize possible calculation strategies, including methods that have been previously adopted and those that have not been utilized before.

\textbf{Input state.} For input state $\mathbf i_t$, We obtain it through a linear projection from input $\mathbf x_t$.



\begin{table*}[t]
\small
    \centering
     \caption{\textbf{Code Mapping of Activation Function}. We perform experiments among the no activation function and seven distinct activation functions.}
     \setlength{\tabcolsep}{2.6mm}
    \begin{tabular}{p{1.4cm}|c|c|c|c|c|c|c|c}
    \toprule
        act\_code & \textbf{\textit{0}} & \textbf{\textit{1}} & \textbf{\textit{2}} & \textbf{\textit{3}} & \textbf{\textit{4}} & \textbf{\textit{5}} & \textbf{\textit{6}} & \textbf{\textit{7}}\\ \hline
        meaning & $x$ & $\mathrm{relu}(x)$ & $\mathrm{sigmoid}(x)$ & $1+\mathrm{elu}(x)$  & $\mathrm{silu}(x)$ & $\mathrm{elu}(x)$ & $\mathrm{relu}^{2}(x)$ & $x^2$ \\ 
    \bottomrule
    \end{tabular}
    \label{table:cmaf}
\end{table*}

\begin{table*}[t]
\small
    \centering
     \caption{\textbf{Code Mapping of Oscillation State}. We use $\color{blue}{\textbf{Blue}}$ to denote data dependent and ${\textbf{Black}}$ to denote data independent and use Einstein Summation notation. For example, ${{k}},{\color{blue}{d}} {{\to}}{\color{blue}{k\ d}}$ means using a data independent $k$ dim vector and a data dependent $\color{blue} d$ dim vector to construct a matrix of size $\color{blue} k\times d$ using outproduct.}
     \setlength{\tabcolsep}{2.6mm}
    \begin{tabular}{p{1.2cm}|c|c|c|c|c|c}
    \toprule
        o\_code & \textbf{\textit{0}} & \textbf{\textit{1}} & \textbf{\textit{2}} & \textbf{\textit{3}} & \textbf{\textit{4}} & \textbf{\textit{5}} \\ \hline
        meaning & ${k\ d}$ & $\color{blue} k, d\to k\ d$ & $\color{blue}d {\overset{broadcast}{\to}}k\ d$ & $\color{blue} k {\overset{broadcast}{\to}}k\ d$ & ${k} {\overset{broadcast}{\to}}{{k}\ d}$ & ${d} {\overset{broadcast}{\to}}{{k}\ d}$\\ \midrule
        o\_code & \textbf{\textit{6}} & \textbf{\textit{7}} & \textbf{\textit{8}} & \textbf{\textit{9}} & \textbf{\textit{10}} & \textbf{\textit{11}}  \\ \hline
        meaning & $k,{{\color{blue}k\ d}} {{\to}}{\color{blue}{k\ d}}$ & $d,{{\color{blue} k\ d}} {{\to}}{{\color{blue} k\ d}}$ & ${k},{\color{blue} d} {{\to}}{\color{blue}  k\ d}$ 
         & ${\color{blue} k},{d} {{\to}}{\color{blue}  k\ d}$  
         & ${ {\mathbf{1}_{k\times d}}}$
         & ${\exp(i {\mathbf \Theta}) \mathbf 1_d^{\top} }$  \\ 
    \bottomrule
    \end{tabular}
    \label{table:cmos}
\end{table*}

\textbf{Parameterization.}
Two parameterization strategies are commonly utilized. The first involves employing a state space model to derive the parameters~\cite{s4}, while the second entails learning the parameters from data through linear projection~\cite{katharopoulos2020transformers}.

\textbf{Data Dependency.}
As indicated in Table~\ref{table:eos}, it is shown that the EOS states of certain methods are dependent on the input state. Specifically, methods with a subscript $t$ in their EOS states exhibit this dependency, while those without a subscript $t$ do not. For data-dependent cases, we assume that each element of the oscillation state belongs to $[0, 1]$ and is calculated using $\mathrm{sigmoid(x)}^{1/\tau}$~\citep{GLA}, where the hyper-parameter $\tau$ controls the oscillation rate. For data-independent cases, we use the method of initialization with Alibi~\citep{alibi}.

\textbf{Construction Methods of Oscillation State.}
The Oscillation state exhibits the most significant variations among the three states. Here, we select 12 distinct scenarios within the Oscillation state, which include the utilization of complex numbers, non-learnable data-independent situations, and the all-ones scenario. We list the following possibilities through the form of Einstein Summation in Table~\ref{table:cmos}. 

\textbf{Activation Function Test.}
In a manner akin to the kernel functions utilized for processing queries and keys in Linear Transformers~\citep{xfmrsarernns}, activation functions are also employed to manipulate the Expand and Shrink states. To assess the efficacy of various activation functions, we conducted tests on several widely-used options. We list them in Table~\ref{table:cmaf}.

\textbf{Tau Test.}
$\tau$ controls the oscillation rate. We also ablate the efficacy of $\tau$ in controlling the oscillation rate. To avoid the oscillation state becoming too small, we used the form $\mathrm{Sigmoid}(x)^{1/\tau}$ following~\citep{GLA}. In this test, we varied the value of $\tau$, with the default value being $\tau=16$ following~\citep{GLA}.

\begin{table}[t]
    \small
    \centering
    \vspace{-3mm}
    \caption{\textbf{Data Dependence Test (left) and $O$ Types Test (right)}. Perplexity values of validation and test split are reported.}
    \setlength{\tabcolsep}{5mm}
        \begin{tabular}{p{2cm}|c|c|c}
        \toprule
        \makecell{Code} & \makecell{ Valid} & \makecell{ Test} & Param. \\ \hline
        $0-0-0-0$ & 29.62 & 30.5 & 43.45 \\  
        $0-0-0-0$ & 28.26 & 29.1 & 43.45 \\  
        $0-0-1-0$ & 28.23 & 29.07 & 43.45 \\ 
        $0-1-0-0$ & 25.89 & 26.61 & 44.04 \\ 
        $0-1-1-0$ & 25.36 & 26.21 & 44.04 \\ 
        $1-0-0-0$ & 27.94 & 28.77 & 43.45 \\ 
        $1-0-1-0$ & 26.12 & 26.65 & 43.45 \\ 
        $1-1-0-0$ & 24.24 & 24.91 & 44.04 \\ 
        $1-1-1-0$ & 24.26 & 24.75 & 44.04 \\ 
        $1-2-1-0$ & 29.27 & 29.77 & 43.45 \\ 
        $1-3-1-0$ & 28.22 & 28.86 & 43.45 \\ 
        $1-4-1-0$ & 27.55 & 28.21 & 44.50 \\ 
        $1-5-1-0$ & 29.01 & 29.82 & 44.51 \\
        $1-6-1-0$ & 25.02 & 25.48 & 44.11 \\ 
        $1-7-1-0$ & 24.82 & 25.17 & 44.04 \\ 
        $1-8-1-0$ & 24.71 & 25.08 & 43.45 \\ 
        $1-9-1-0$ & 24.23 & 24.54 & 43.46 \\
        $1-10-1-0$ & 29.91 & 30.31 & 43.45 \\
        $1-11-1-0$ & 28.6 & 29.36 & 44.11 \\
        \bottomrule
    \end{tabular}
    \vspace{-4mm}
    \label{table:ppl_exp1}
\end{table}

\subsection{Model Code}
For the ease of representing the model variants, we use model codes to classify different variants. Given the following variables: parameterization type, expand type, oscillation type, shrink type, activation function type, we use the format `e-o-s-a` to denote the model code. We use a separate code 0 to denote SSM parameterization methods like Mamba, and S4, whereas other codes represent linear parameterization.

\begin{table}
    \small
    \centering
    \caption{\textbf{Parameterization Test}. Perplexity values of validation and test split are reported.}
    \setlength{\tabcolsep}{3.3mm}
    \begin{tabular}{p{1.9cm}|c|c|c}
    \toprule
        Code & \makecell{ Valid} & \makecell{ Test} & Param. \\ \hline
        $0$ & 27.41 & 27.99 & 43.78 \\ 
        $1-1-1-0$ & 24.26 & 24.75 & 44.04 \\
    \bottomrule
    \end{tabular}
    \label{table:parameterization_test}
\end{table}

\begin{table*}[t]
    \small
    \centering
    \caption{\textbf{Activation Function Test(left) and Tau Value Test (right)}. Perplexity values of validation and test split are reported.}
    \setlength{\tabcolsep}{5.5mm}
    \begin{minipage}{0.55\hsize}
        \begin{tabular}{p{2cm}|c|c|c}
        \toprule
        Code & \makecell{ Valid} & \makecell{ Test} & Param. \\ \hline
        $1-1-1-0$ & 24.26 & 24.75 & 44.04 \\ 
        $1-1-1-0$ & 23.78 & 24.97 &46.07   \\ 
        $1-1-1-1$ & 22.85 & 23.3 & 46.07 \\ 
        $1-1-1-2$ & 23.04 & 23.61 & 46.07 \\
        $1-1-1-3$ & 22.92 & 23.47 & 46.07 \\ 
        $1-1-1-4$ & 22.73 & 23.22 & 46.07 \\
        $1-1-1-5$ & 23.31 & 23.7 & 46.07 \\
        $1-1-1-6$ & 22.8 & 23.32 & 46.07 \\ 
        $1-1-1-7$ & 23.77 & 23.91 & 46.07 \\
        \bottomrule
    \end{tabular}
    \end{minipage}%
    \begin{minipage}{0.5\hsize}
        \begin{tabular}{p{0.8cm}|c|c|c}
    \toprule
        Tau & \makecell{ Valid} & \makecell{ Test} & Param. \\ \hline
        1 & 25.77 & 26.24 & 44.04 \\ [3pt] 
        2 & 24.8 & 25.29 & 44.04 \\  [3pt]
        4 & 23.98 & 24.53 & 44.04 \\ [3pt]
        8 & 23.53 & 24.0 & 44.04 \\  [3pt]
        16 & 24.26 & 24.75 & 44.04 \\ [3pt]
        32 & 25.1 & 25.36 & 44.04 \\  [3pt]
        64 & 24.58 & 24.96 & 44.04 \\ [3pt]
    \bottomrule
    \end{tabular}
    \end{minipage}%
    \label{table:ppl_exp2}
\end{table*}

\begin{figure*}[t]
    \centering
    \vspace{-4mm}
    \includegraphics[width=1\textwidth]{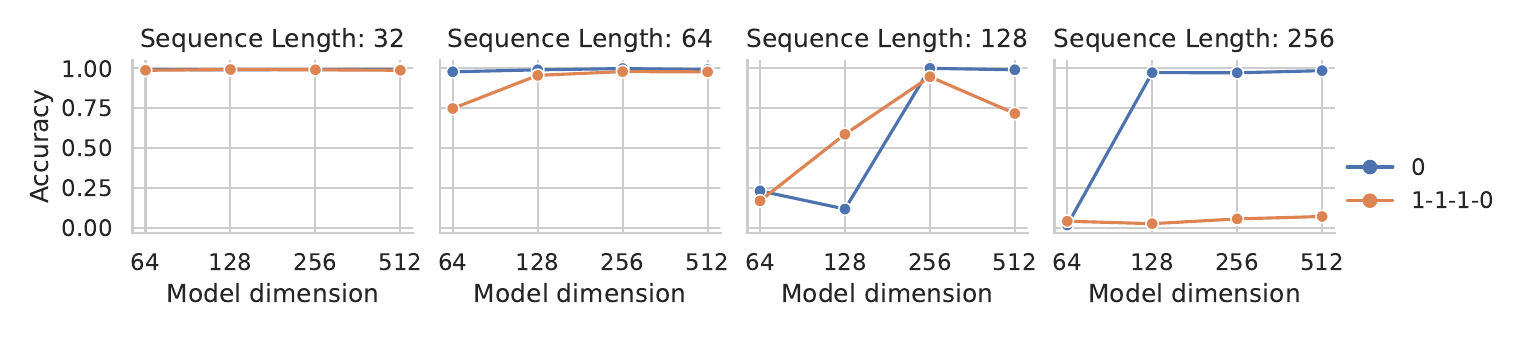}
    \vspace{-10mm}
    \caption{\textbf{Parameterization Test}. The accuracy for various sequence lengths and model dimensions are reported.}
    \vspace{-4mm}
    \label{fig:parameter}
\end{figure*}
\begin{figure*}[t]
    \centering
    \includegraphics[width=1\textwidth, trim={0 6mm 0 0},clip]{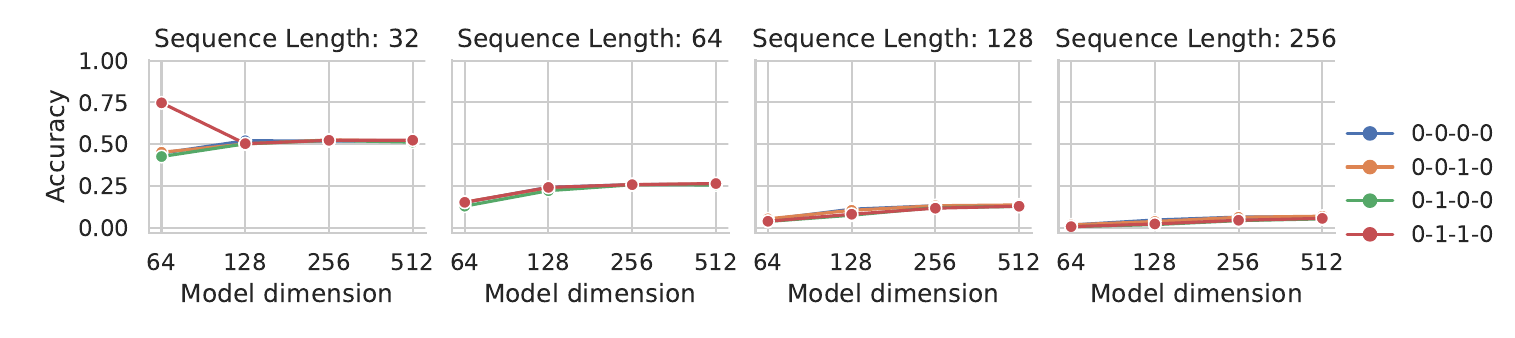}
    \includegraphics[width=1\textwidth]{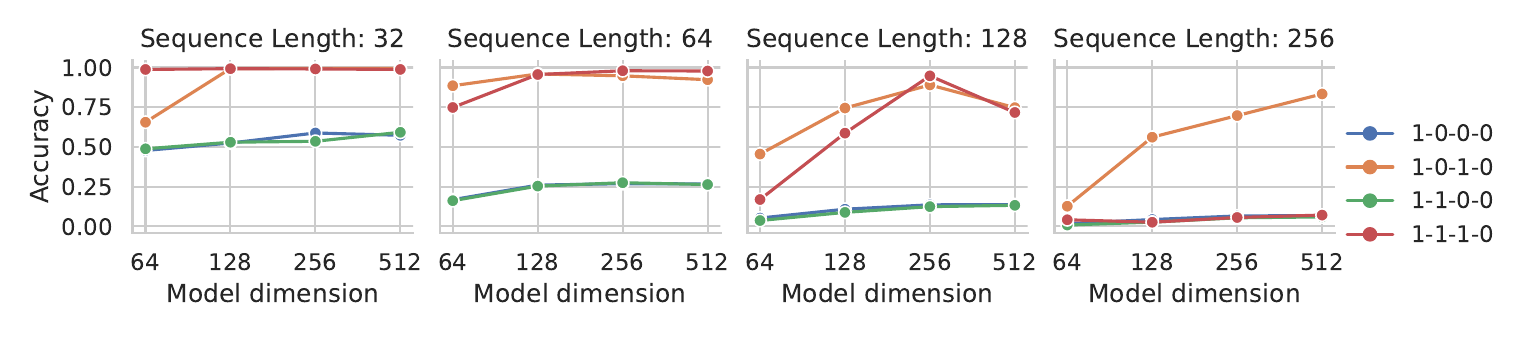}
    \vspace{-10mm}
    \caption{\textbf{Data Dependence Test}. The accuracy for various sequence lengths and model dimensions are reported.}
    \vspace{-4mm}
    \label{fig:ddt}
\end{figure*}
\begin{figure*}[t]
    \centering
    \vspace{-4mm}
    \includegraphics[width=1\textwidth, trim={0 6mm 0 0},clip]{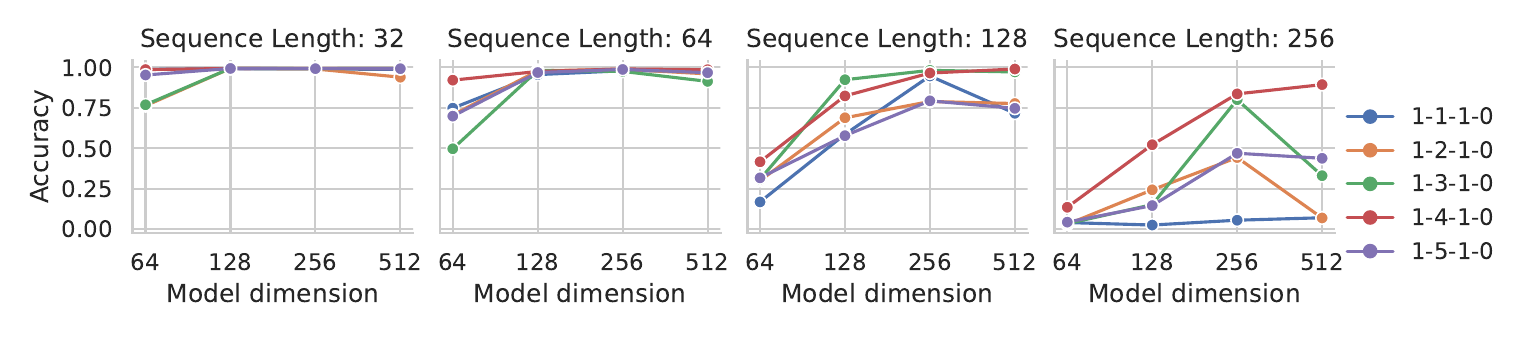}
    \includegraphics[width=1\textwidth]{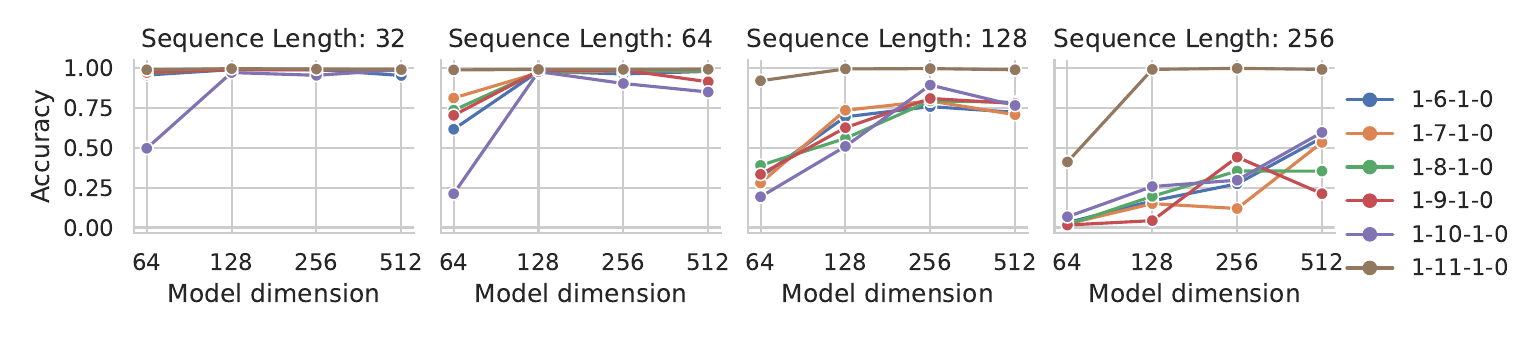}
    \vspace{-10mm}
    \caption{\textbf{O Types Test}. The accuracy for various sequence lengths and model dimensions are reported.}
    \vspace{-4mm}
    \label{fig:otype_test}
\end{figure*}

\begin{figure*}[h]
    \centering
    \vspace{-4mm}
    \includegraphics[width=1\textwidth, trim={0 6mm 0 0},clip]{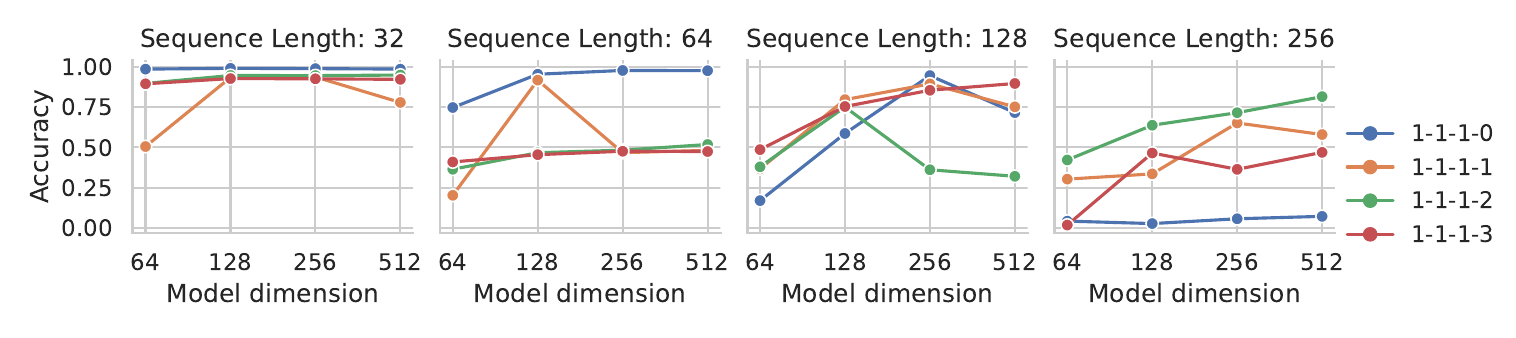}
    \includegraphics[width=1\textwidth]{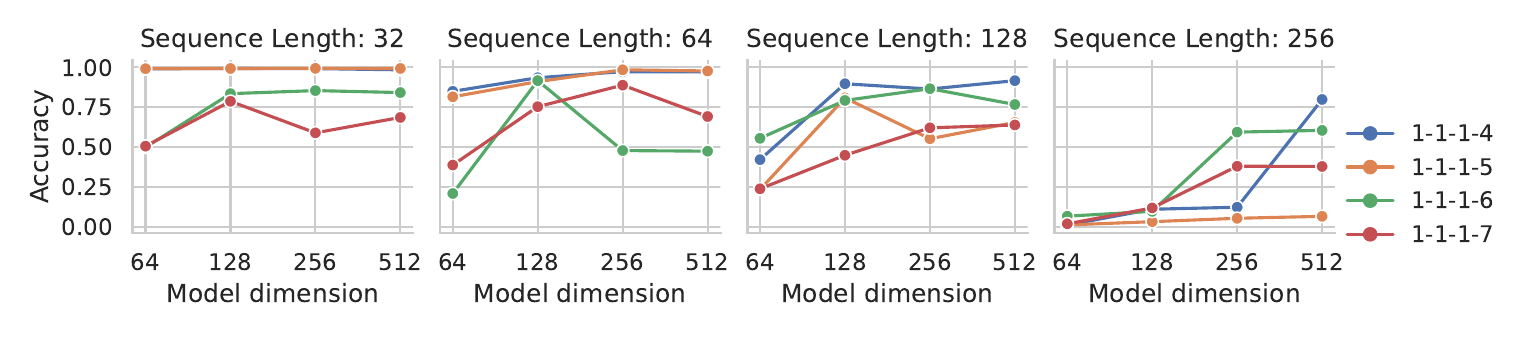}
    \vspace{-10mm}
    \caption{\textbf{Activation Functions Test}. The accuracy for various sequence lengths and model dimensions are reported.}
    \vspace{-4mm}
    \label{fig:act_test}
\end{figure*}
In this context, $e$ and $s$ take values of 0 or 1, representing the expand type and shrink type, where 0 indicates data independence and 1 indicates data dependence. The variable $o$ ranges from 0 to 11 and denotes the method of constructing the oscillation state, with the complete table detailed in Table \ref{table:cmos}. Similarly, the variable $a$ ranges from 0 to 7 and indicates the type of activation function, with the complete table shown in Table \ref{table:cmaf}. For instance, a setting of ``0-0-0-0'' signifies the use of linear parameterization, with all states being data independent and no activation function being utilized.

\section{Experiments}
We perform extensive experiments to evaluate the model variants in language modeling performance and long-context recall capabilities on WikiText-103~\citep{merity2017pointer} dataset and the multi-query associative recall (MQAR) task~\citep{zoology}. The experiment setups are delineated below.

For the convergence experiments on WikiText-103, we adhered to the configuration outlined in prior studies~\cite{qin2023toeplitz}. Specifically, we employed a batch size of 128, and a sequence length of 512, and ran the experiments for 50,000 iterations. The learning rate was set to 5e-4, with complete hyperparameters listed in Appendix ~\ref{configuration}.

For the MQAR task, we standardized the expand ratio $k$ for all models to 128. We conducted a grid search for the learning rate, spanning values from \{1e-5, 5e-5, 1e-4, 5e-4, 1e-3, 5e-3, 1e-2\}, and reported the best results for each model. Additionally, we tested various sequence lengths {32, 64, 128, 256}, and feature dimensions {64, 128, 256, 512} to determine the optimal configuration.

\subsection{Experiments on Wikitext-103}
\paragraph{Parameterization Test.}

We carried out a comparison between SSM parameterization and naive parameterization, with the findings presented in Table ~\ref{table:parameterization_test}. The analysis revealed that SSM parameterization showed inferior performance compared to naive parameterization on the Wikitext dataset.

\vspace{0mm}
\paragraph{Data Dependence Test.}
We proceeded to assess the extent of dependency on data-dependent states, as shown in Table ~\ref{table:ppl_exp1}. Analysis of the Wikitext data indicates that when all EOS states ($e, o, s$) are non-data-dependent, the performance is notably subpar, even worse than the scenario where the $o$ state is not learnable. Introducing data dependence on just one of these elements, particularly on the $o$ state ($0-1-0-0$), demonstrates the most significant effectiveness. When data dependence is introduced on two elements, the combination of $o$ and $s$ ($1-1-0-0$) proves to be the most effective, yielding results comparable to scenarios where $e$, $o$, and $s$ are all data-dependent.


\vspace{0mm}
\paragraph{O Types Test.}
In Table~\ref{table:ppl_exp1}, we evaluated various strategies for constructing the $o$ state. A method we labeled as $1-9-1-0$, yielded the most favorable outcomes. This approach outperformed the $1-1-1-0$ configuration. On the other hand, methods that relied on a naive $\mathbf o$ state (such as linear attention~\citep{xfmrsarernns}) demonstrated inferior performance.


\vspace{0mm}
\paragraph{Activation Functions Test.}
Then we examined the influence of activation functions on the outcomes, results are presented in Table~\ref{table:ppl_exp2}. It's clear that employing activation functions yields considerable benefits, with $\mathrm{silu}$ and $\mathrm{relu}^2$ demonstrating the most favorable performance.


\vspace{0mm}
\paragraph{Tau Values Test.}
Finally we explored the determination of tau values for the $1-1-1-0$ configuration, results are also listed in Table~\ref{table:ppl_exp2}. On the WikiTest-103 task, selecting tau=8 resulted in the most favorable outcomes.


\vspace{-2mm}
\subsection{Experiments on MQAR}
\vspace{0mm}
\paragraph{Parameterization Test.}
In Fig.\ref{fig:parameter}, a comparative analysis was carried out between SSM parameterization and naive parameterization, illustrating that naive parameterization achieves inferior performance in the MQAR task.

\vspace{0mm}
\paragraph{Data Dependence Test.}
We proceeded to evaluate the reliance on data-dependent states, as shown in Fig.~\ref{fig:ddt}. In the MQAR task, the majority of configurations resulted in subpar performance, with the exception of the $1-0-1$ and $1-1-1$ configurations, which showed satisfactory results for sequence lengths up to 128. Nonetheless, when the sequence length was increased to 256, solely the $1-0-1-0$ configuration managed to converge successfully.

\vspace{0mm}
\paragraph{O Types Test.}
In Fig.~\ref{fig:otype_test}, we conducted a comparison of various methods for constructing $o$ state. Within the context of the MQAR task, the performance of the complex variant stood out as particularly favorable, surpassing other variants significantly.

\vspace{0mm}
\paragraph{Activation Functions Test.}
In Figure~\ref{fig:act_test}, we explored the impact of activation functions. Employing activation functions resulted in a significant improvement in performance, particularly with longer sequence lengths. Additionally, the $sigmoid (1-1-1-2)$ activation function surpassed the performance of other activation functions.

\begin{figure*}[htpb]
    \centering
    \vspace{-4mm}
    \includegraphics[width=1\textwidth,trim={0 6mm 0 0},clip]{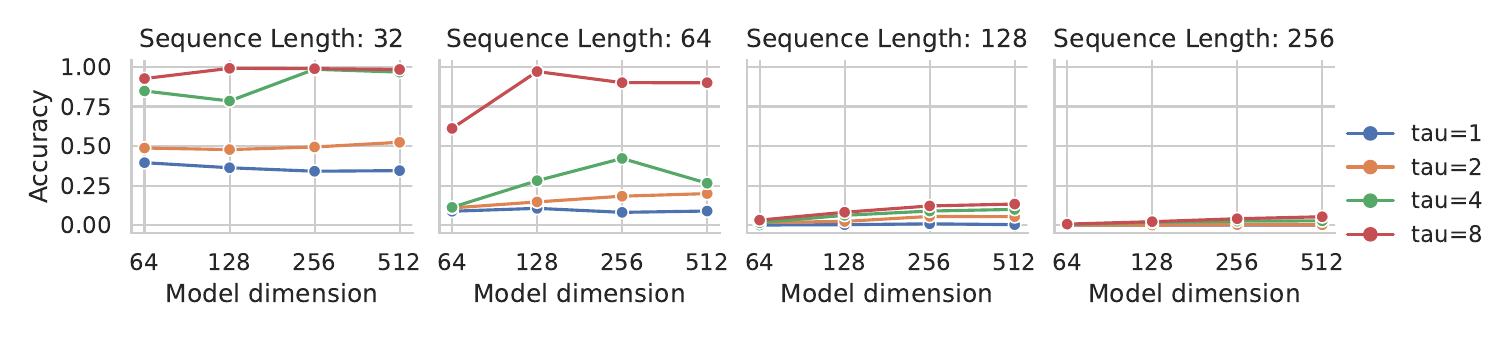}
    \includegraphics[width=1\textwidth]{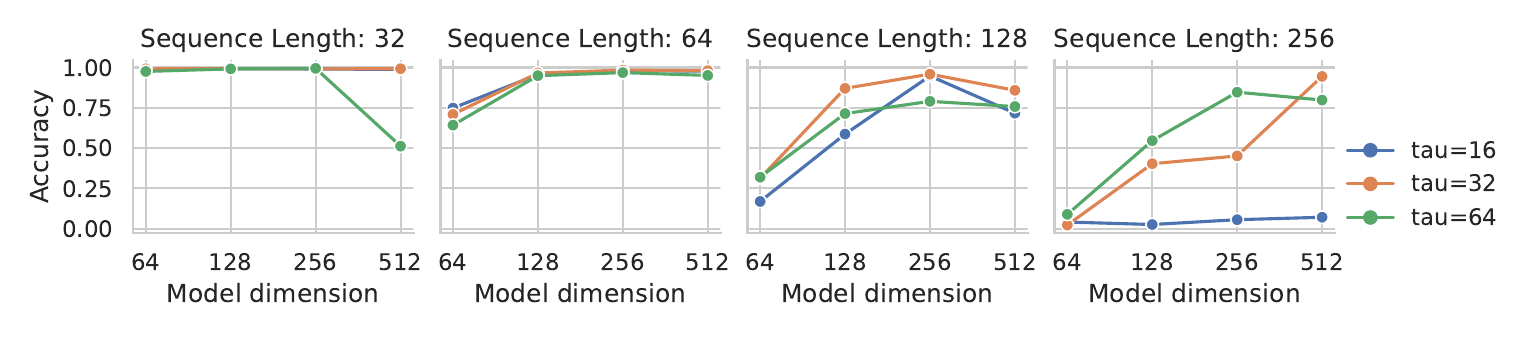}
    \vspace{-10mm}
    \caption{\textbf{Tau Values Test}. The accuracy for various sequence lengths and model dimensions are reported.}
    \vspace{-4mm}
    \label{fig:tau_test}
\end{figure*}

\vspace{0mm}
\paragraph{Tau Values Test.}
We delved into establishing the tau values for the $1-1-1-0$ configuration. As shown in Fig.~\ref{fig:tau_test}, selecting a larger $\tau$ , (thus a large $o$ state) led to enhanced outcomes for MQAR task, which aligns with the capability for long document retrieval.

\section{Conclusion}
\vspace{0mm}
This paper introduced the Linear Complexity Sequence Model (LCSM), a unified framework that synergizes various linear complexity sequence modeling techniques. Our exploration of the LCSM's three integral stages—Expand, Oscillation, and Shrink (EOS)—and the role of the Memory state has unveiled the critical impact of data dependency and parameterization on model performance. The LCSM framework not only enhances our understanding of the operational dynamics of linear complexity models but also opens avenues for future research to further refine and adapt these models for improved efficiency and effectiveness across a broader spectrum of sequence modeling challenges.

\bibliography{lcsm}
\bibliographystyle{icml2024}
\section{Appendix}

\subsection{Backward Pass}
In the main text, we only defined the forward pass of LCSM. Here, we define the backward pass of LCSM. For convenience, we will refer to the case where ($\psi=\odot$) as Type1, and the case where ($\psi=\times $) as Type2.

 \textbf{Type1.}
The forward pass of Type 1 is:
\begin{equation*}
\mathbf m_{t}=\mathbf f_t \odot \mathbf m_{t-1} + \mathbf e_t \mathbf i_t^\top,\\
\mathbf y_t =\mathbf m_t^{\top} \mathbf s_t .
\end{equation*}
So the backward pass can be computed as follows:
\begin{equation}
\begin{aligned}
\mathbf {ds}_t& = \mathbf m_t \mathbf {dy_t} \in \mathbb R^{k},\\
\mathbf {dm}_{t-1}& =\mathbf f_t \odot \mathbf {dm}_{t} + \mathbf s_{t-1}  \mathbf {dy}_{t-1} ^{\top}\in \mathbb R^{k\times d}, \\
\mathbf {df}_{t}& =\mathbf {dm}_{t} \odot \mathbf m_t \in \mathbb R^{k\times d}, \\
\mathbf {de}_{t}& = \mathbf {dm}_{t} \mathbf i_t \in \mathbb R^{k}, \\
\mathbf {di}_{t}& = \mathbf {dm}_{t}^{\top} \mathbf e_t \in \mathbb R^{d}. 
\end{aligned}
\end{equation}

\textbf{Type2.} The forward pass of Type2 is:
$$
\mathbf m_{t}=\mathbf f_t\mathbf m_{t-1} + \mathbf e_t \mathbf i_t^\top,\\
\mathbf y_t =\mathbf m_t^{\top} \mathbf s_t.
$$
So the backward pass can be computed as follows:
\begin{equation}
\begin{aligned}
\mathbf {ds}_t &= \mathbf m_t \mathbf {dy_t} \in \mathbb R^{k},\\
\mathbf {dm}_{t-1}&=\mathbf f_t \mathbf {dm}_{t} + \mathbf s_{t-1}  \mathbf {dy}_{t-1} ^{\top}\in \mathbb R^{k\times d}, \\
\mathbf {df}_{t}&=  \mathbf {dm}_{t} \mathbf m_t^{\top} \in \mathbb R^{k\times k}, \\
\mathbf {de}_{t}&= \mathbf {dm}_{t} \mathbf i_t \in \mathbb R^{k}, \\
\mathbf {di}_{t}&= \mathbf {dm}_{t}^{\top} \mathbf e_t \in \mathbb R^{d}. 
\end{aligned}
\end{equation}

\subsection{More discussion about $g=\times$}
\label{no_diag}
In the main text, we primarily discuss the case of $g=\odot$ and assume that $\mathbf{o}_t$ is diagonalizable when $g=\times$. In this section, we will discuss the more general case of $g=\times$. that is, the non-diagonalizable case, which we will list in Table~\ref{table:matrix}.

\textbf{FWP}~\citep{2102.11174}
\begin{equation}
\begin{aligned}
\mathbf W_t
&=\mathbf W_{t-1} + \beta_t \mathbf k_t (\mathbf v_t - \bar{\mathbf v}_t)^\top \\
&=\mathbf W_{t-1} + \beta_t \mathbf k_t (\mathbf v_t -\mathbf W_{t-1}^\top  \mathbf k_t)^\top \\
&=(\mathbf I_k -\beta_t \mathbf k_t \mathbf k_t^\top) \mathbf W_{t-1} + \beta_t \mathbf k_t \mathbf v_t^\top \\
\mathbf y_t& = \mathbf W_t^\top \mathbf q_t.
\end{aligned}
\end{equation}
Based on the aforementioned forms, we propose several new forms for future research in Table~\ref{table:matrix},

\begin{table*}[!ht]
    \small
    \centering
    \caption{The checklist when $\mathbf o_t$ is not diagonalizable, we also give M1, M2, M3 for future research.}
    \setlength{\tabcolsep}{3mm}
    \begin{tabular}{l|c|c|c|c|c|c}
    \toprule
Method           & Input                     & Expand                          & Oscillation                                                 & Shrink                    & Memory Size $k\times d$  & $g$ \\ \midrule
FWP & $ \mathbf v_t$ & $  \beta_t \mathbf k_t$     & $\mathbf I_k -\beta_t \mathbf k_t  \mathbf k_t^\top $                       & $ \mathbf q_t$  & $k\times d$  & $\times $   \\ 
M1 & $ \mathbf v_t$ & $  \mathbf k_t$     & $\mathbf I_k - \mathbf k_t  \mathbf k_t^\top $                       & $ \mathbf q_t$  & $k\times d$  & $\times $   \\ 
M2 & $ \mathbf v_t$ & $  \mathbf k_t$     & $\mathbf I_k - \mathbf k_t  \bar {\mathbf k}_t^\top $                       & $ \mathbf q_t$  & $k\times d$  & $\times $   \\ 
M3 & $ \mathbf v_t$ & $  \mathbf k_t$     & $\mathbf I_k - \mathrm{Diag}\{\mathbf k_t \}   $                       & $ \mathbf q_t$  & $k\times d$  & $\times $   \\ 
        \bottomrule
    \end{tabular}
    \label{table:matrix}
\end{table*}

\subsection{Prove the equivalent expression of Cosformer and LRPE.}
\label{proof}

\textbf{Cosformer.}
\begin{equation}
\begin{aligned}
{[\mathbf {kv}]_t } &= \sum_{s=1}^t \Lambda^{t-s}\mathbf k_s\mathbf  v_s^\top \\
\mathbf y_t&= \mathrm{Rel}\{[\mathbf {kv}_t] \}^{\top} \mathbf q_t \\
&= \mathrm{Rel}\left\{
\sum_{s=1}^t \Lambda^{t-s}\mathbf v_s \mathbf  k_s ^\top\mathbf q_t
\right\} \\
&=
\sum_{s=1}^t \mathbf v_s \mathbf  k_s ^\top \bar \Lambda_{t-s}\mathbf q_t
 \\
 \bar  \Lambda_{t-s}&= \mathrm{diag}\{\cos((t-s)\theta_1),\ldots, \cos((t-s)\theta_k) \}.
\end{aligned}
\end{equation}

\textbf{LRPE.}
\begin{equation}
\begin{aligned}
{[\mathbf {kv}]_t }&= \sum_{s=1}^t \exp(i (t-s) \theta)\mathbf k_s\mathbf  v_s^\top \\
\mathbf y_t&= \mathrm{Rel}\{[\mathbf {kv}]_t \}^{\top} \mathbf q_t \\
&= \mathrm{Rel}\left\{
\sum_{s=1}^t \exp(i (t-s) \theta)\mathbf v_s \mathbf  k_s ^\top\mathbf q_t
\right\} \\
&=
\sum_{s=1}^t \cos((t-s) \theta)\mathbf v_s \mathbf  k_s ^\top\mathbf q_t.
\end{aligned}
\end{equation}

\section{Configurations}
\label{configuration}
\begin{table}[h]
\small
\center
\setlength{\tabcolsep}{4mm}
{
\caption{Detailed training configurations used in for Language Modeling task. ``Total batch size'' means $\mathrm{batch\_per\_gpu} \times \mathrm{update\_freq} \times \mathrm{num\_gpus}$.}
\label{Tconfiguration}
\begin{tabular}{l|l}
\toprule
 & LM   \\
\midrule
Data                                              & WikiText-103                          \\
Tokenizer method & BPE  \\
Src Vocab size & 50265  \\
Sequence length    & 512          \\
Total batch size & 128                                              \\
Number of updates/epochs                            & 50k updates     \\
Warmup steps/epochs    & 4k steps       \\
Peak learning rate   & 5e-4                            \\
Learning rate scheduler  & Inverse sqrt                     \\
Optimizer   & Adam                                      \\
Adam $\epsilon$   & 1e-8                             \\
Adam $(\beta_1,\beta_2)$                          & (0.9, 0.98)             \\
Weight decay       & 0.1 \\                                       
Gradient clipping                                            &  -                                             \\
\bottomrule
\end{tabular}}

\end{table}






\begin{table}[h]
    \centering
    \caption{\textbf{Additional Activation Functions Test}. Perplexity values of validation and test split are reported.}
    \setlength{\tabcolsep}{3mm}
    \begin{tabular}{p{2.5cm}|c|c|c}
    \toprule
        Code & PPL (Valid) & PPL (Test) & Paramerters \\ \hline
        $1-10-1-0$ & 29.91 & 30.31 & 43.45 \\ 
        $1-10-1-1$ & 27.02 & 27.44 & 43.45 \\
        $1-10-1-2$ & 26.77 & 27.37 & 43.45 \\ 
        $1-10-1-3$ & 26.82 & 27.6 & 43.45 \\
        $1-10-1-4$ & 27.28 & 27.58 & 43.45 \\ 
        $1-10-1-5$ & 28.64 & 28.98 & 43.45 \\
        $1-10-1-6$ & 26.68 & 27.18 & 43.45 \\
        $1-10-1-7$ & 28.08 & 28.63 & 43.45 \\ 
    \bottomrule
    \end{tabular}
\end{table}

\end{document}